\pdfoutput=1

\documentclass{article}

\usepackage[preprint]{neurips_2025}
\makeatletter
\renewcommand{\@noticestring}{2nd Workshop on Principles of Generative Modeling (PriGM), NeurIPS 2026.}
\makeatother

\usepackage[utf8]{inputenc}
\usepackage[T1]{fontenc}
\usepackage{url}
\usepackage{booktabs}
\usepackage{amsfonts}
\usepackage{amsmath}
\usepackage{amssymb}
\usepackage{amsthm}
\usepackage{microtype}
\usepackage{graphicx}
\usepackage{subcaption}
\usepackage{natbib}
\usepackage{hyperref}

\newcommand{\Ldream}{\mathcal{L}_{\mathrm{dream}}}
\newcommand{\Lreg}{\mathcal{L}_{\mathrm{reg}}}

\newcommand{\KL}{D_{\mathrm{KL}}}

\theoremstyle{plain}

\theoremstyle{definition}

\theoremstyle{remark}

\title{Mitigating Model Collapse through Dreaming Learning}

\author{%
  Matteo Benati\\
  DIAG, Sapienza University of Rome, Italy\\
  \texttt{matteo.benati@uniroma1.it}
  \And
  Alessandro Londei\\
  Sony CSL Rome, CREF, Rome, Italy
  \And
  Denise Lanzieri\\
  Sony CSL Rome, CREF, Rome, Italy
  \AND
  Raffaello Mastromarino\\
  Copenhagen Center for Social Data Science (SODAS),\\
  University of Copenhagen, Denmark
  \And
  Vittorio Loreto\\
  Department of Physics, Sapienza University of Rome, Italy\\
  Sony CSL Rome, CREF, Rome, Italy\\
  Complexity Science Hub, Vienna, Austria
}

\begin{document}

\maketitle

\begin{abstract}
Training generative models on synthetic data causes loss of diversity and
a decrease in performance, a phenomenon known as model collapse. We study
this degeneration spectrally in a controlled Markov setting. The
eigenvalue spectrum of the learned transition operator offers a direct
view of the process, since the subdominant eigenvalues set how fast the
chain mixes and therefore how much of the state space it continues to
explore. We find that sampling at temperature $T_s < 1$ acts on the
operator like repeated sharpening, pushing it toward a deterministic
system: the subdominant eigenvalues approach the unit circle and the
spectral gap narrows. When $T_s < 1$, even data accumulation does not
appear to arrest this drift. We then analyze Dreaming Learning
\citep{londei2024}, a regulariser that, in parallel with regular training,
adds a loss term on the model's own samples drawn at a raised temperature
$T_d > 1$. Because high-temperature samples flatten the predicted
distribution, training on them keeps the entries of the learned operator
from becoming vanishingly small, which in turn keeps the eigenvalues
bounded away from the unit circle. Measurements of the spectra over $5000$
generations are consistent with this picture, and the mitigation effect
persists in more complex settings such as LSTM language models and a
124M-parameter GPT-2.
\end{abstract}

\section{Introduction}
As synthetic data is an ever-increasing percentage of online content, generative models are increasingly retrained on data they themselves produced, without being able to distinguish human-made content from machine-made one. This phenomenon lowers diversity and drives learned distributions toward low-entropy distributions, an effect observed across GANs, diffusion models and language models \citep{shumailov2024,dohmatob2024a,liu2022,yoon2024,guo2024}. The leading mitigation technique is \emph{accumulation}: retaining a fraction of real data alongside synthetic data \citep{gerstgrasser2024,seddik2024}. Accumulation bounds test error, but existing analyses fix the sampling temperature used in language modeling experiments at $T_s = 1$, almost never used during inference \citep{renze2024}, and no consensus exists on real-to-synthetic ratios \citep{dohmatob2024b}. Furthermore, datasets cannot always be distinguished upfront, risking sub-optimal mitigation.

We approach the problem through the spectrum of the learned transition operator rather than through test error alone. Test error reports how far the model has moved from the target, but it says little about which structural property of the generative process is being lost. The object we track instead is the conditional structure of the process itself. In a Markov setting this structure is a transition operator, or transition matrix, and its eigenvalues give information that a scalar loss does not: the subdominant eigenvalues control how fast the chain mixes and therefore how much of the state space keeps being visited. Watching them evolve across generations gives a clearer picture of collapse as it happens rather than just a degradation in performance. Framed this way, self-consuming retraining is a question about which structural
properties of a generative model survive post-training on its own outputs, and the
sampling temperature is an inference-time choice that turns out to govern the answer. Our contributions are:

\begin{itemize}\itemsep2pt \parskip0pt
\item \textbf{A spectral view of collapse.} Training on synthetic data sampled at $T_s<1$ acts on the operator as a sharpening map whose fixed points are deterministic transition matrices. The subdominant eigenvalues approach $|\lambda_{i>1}| \to 1$ and the spectral gap narrows.
\item \textbf{A mechanism that mitigates it.} Adding a loss term on
synthetic sequences drawn at $T_d > 1$ flattens the sampled distribution
and keeps the entries of the operator from approaching zero. This in turn
keeps $|\lambda_{i>1}|$ away from $1$ as $t$ grows.
\item \textbf{Empirical measurements} across three scales: feed-forward model on Markov chains, LSTM language models and GPT-2 LLM synthetic finetuning (124M), following the protocol of \citet{shumailov2024}.
\end{itemize}

\section{Collapse and the spectral gap}
\label{sec:spectral}
The second-largest eigenvalue in modulus, $|\lambda_2|$, sets the
asymptotic rate at which a chain converges to its stationary distribution \citep{levin2017}, so the spectrum of the transition matrix says how fast the process forgets its current state.
Recursive training on self-generated data degrades a model even when
sampling is unbiased, because finite samples under-represent the tails of
each conditional \citep{shumailov2024}. Sampling below unit temperature
adds a second route: outputs are biased toward the mode of each
conditional, and retraining on them makes the conditionals more peaked in
turn. On the transition matrix one round of this is the row-wise
sharpening operator
\begin{equation}
\label{eq:sharpen}
\mathcal{S}_T(M)_{ij} = \frac{M_{ij}^{1/T}}{\sum_k M_{ik}^{1/T}},
\end{equation}
with $T = T_s$. Applying \eqref{eq:sharpen} repeatedly concentrates each
row on its largest entry, so the matrix drifts toward determinism. Our
observations at $T_s = 1$ isolate the tail-truncation route alone.
A deterministic row-stochastic matrix is the transition matrix of a map
from the state space to itself, whose recurrent part is a union of cycles.
A cycle of length $L > 1$ contributes the $L$-th roots of unity, so the
subdominant eigenvalues approach the unit circle and the spectral gap
narrows.\\
Sampling at $T = T_d > 1$ acts in the opposite direction, flattening each row and keeping entries away from zero. A row-stochastic matrix with all entries at least $c$ has Dobrushin ergodicity coefficient at most $1 - nc$, which bounds every eigenvalue other than the Perron root \citep{levin2017}. Dreaming Learning supplies this floor: high-temperature samples spread mass across all states, ensuring nonzero support in the network's learned conditionals. The spectrum directly validates the mechanism.

\section{Dreaming Learning}
\label{sec:dreaming}
Dreaming Learning \citep{londei2024} augments standard training with an
auxiliary dreaming phase, in which the network generates synthetic sequences during training and
then learns from them. It requires that the network be probabilistic and
that its output be mappable back into its input space. We use it as a
regularization term.
At each step the model generates a synthetic batch $\tilde{x}^{(i)}_{1:\ell}$,
$i = 1,\dots,k$, of length $\ell$, by recursively sampling from its own Softmax
outputs tempered at $T_d$, which is~\eqref{eq:sharpen} applied to the model's
conditionals with $T = T_d$. The dreaming loss is the cross-entropy of the model
on that batch,
\begin{equation}
\Ldream = -\frac{1}{k(\ell-1)} \sum_{i=1}^{k} \sum_{t=1}^{\ell-1}
\log P\!\left( \tilde{x}^{(i)}_{t+1} \mid \tilde{x}^{(i)}_{1:t} ; \Omega \right),
\qquad
\mathcal{L} = \Lreg + \alpha \Ldream ,
\label{eq:dreaming}
\end{equation}
where $\mathcal{L}_{\mathrm{reg}}$ is the usual next-token cross-entropy
on real data and $\alpha$ controls the strength of the regularizer.

Unlike the original formulation \citep{londei2024}, which targets adaptation to regime
shifts, our implementation is epoch-based and uses the dreaming phase to
preserve statistical features across repeated retraining. Summing the two losses encourages the model both to fit real data and to
explore its own Adjacent Possible \citep{adjacentpossible, Tria2014}, which
has been associated with robustness in data statistics
\citep{londei2024}. Related work reaches a similar conclusion from the
data side, where selection criteria that reduce diversity in the training
corpus degrade general performance
\citep{fan2025combattingdimensionalcollapsellm}.
\section{Experiments}
\label{sec:exp}

\subsection{Markov chains}
\label{sec:markov}

The behaviour described in Section~\ref{sec:spectral} can be observed
directly in a controlled chain. We fix a Markov transition matrix $M_0$ with $n=10$ states and normalised stationary distribution's entropy of $0.8$, and draw a baseline dataset of $20{,}000$ states from it. Each generation runs the following loop: sample $N_g = 5000$ new symbols at $T_s = 0.7$ from the current model, append them to the dataset so that data accumulates rather than being replaced, and retrain the model on the accumulated data.

The model is a small feed-forward network that predicts the next symbol from the current one, so the transition matrix $M_t$ can be read off its outputs at every generation. Dreaming enters only through the training objective. In the dreaming condition the network is trained with the total loss of \eqref{eq:dreaming}, where the dreaming term is computed on $N_d = 5000$ symbols the network generates from its own tempered outputs at $T_d$. In the vanilla condition $\alpha = 0$ and the second term is absent. All results are over 10 seeds. Details are in Appendix~\ref{app:markov}.

\begin{figure}
\centering
\makebox[0pt]{\includegraphics[width=1.0\linewidth]{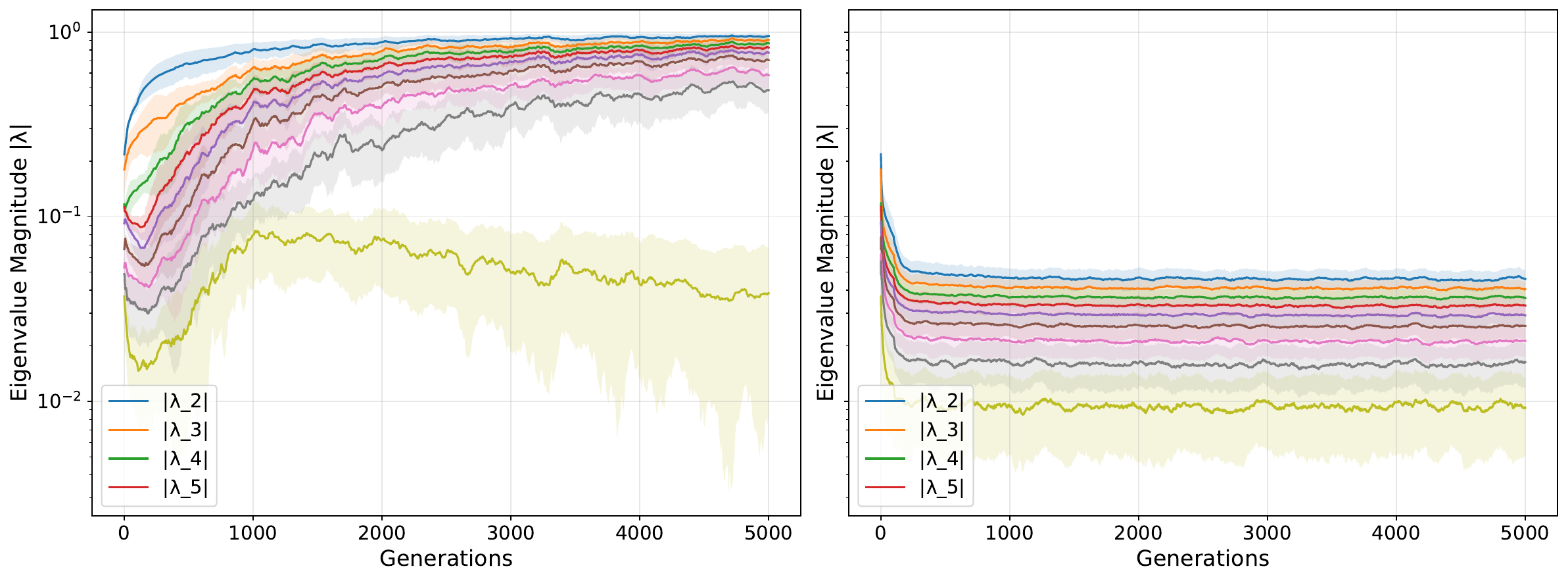}}
\caption{Eigenvalue magnitudes $|\lambda_2|,\dots,|\lambda_5|$ of the
reconstructed transition matrix across 5000 generations and 10 seeds,
moving average ($w{=}100$), shaded region represents 25th to 75th
percentile. Dashed lines mark the true
spectrum of $M_0$. \textbf{Left:} vanilla recursion. The subdominant
eigenvalues climb toward $1$ and the spectral gap narrows.
\textbf{Right:} with Dreaming Learning ($T_d{=}1.5$, $\alpha{=}0.8$)
the spectrum is flat in $t$. The dreaming spectrum stabilises
\emph{below} the true one, indicating a residual bias toward faster
mixing that is controlled by $T_d$ and $\alpha$.}
\vspace{-0.3cm}
\label{fig:spectrum}
\end{figure}

Because $M_t$ is available at every generation, its spectrum can be measured directly. Figure~\ref{fig:spectrum} tracks $|\lambda_2|,\dots,|\lambda_5|$ over $5000$ generations. Under the vanilla recursion the subdominant eigenvalues drift upward toward the unit circle and the spectral gap narrows, which is the behaviour anticipated for repeated sharpening at $T_s < 1$. With dreaming enabled the same eigenvalues remain approximately constant in $t$. The spectrum settles below the true one, so the reconstructed chain mixes somewhat faster than $M_0$, but it does not drift over the $5000$ generations we observe. The size of this offset depends on $T_d$ and $\alpha$.

Figure~\ref{fig:kl} shows the KL divergence between true and reconstructed stationary distributions after 5000 generations. The divergence improves by $> 2$ orders of magnitude at $T_d = 1.5, \alpha = 0.8$, consistent with strong dreaming keeping transition probabilities away from zero.

\subsection{Language models}

\begin{figure}[t]
\centering
\vspace{-0.3cm}
  \includegraphics[width=0.8\linewidth, trim={0 0 0 27}, clip]{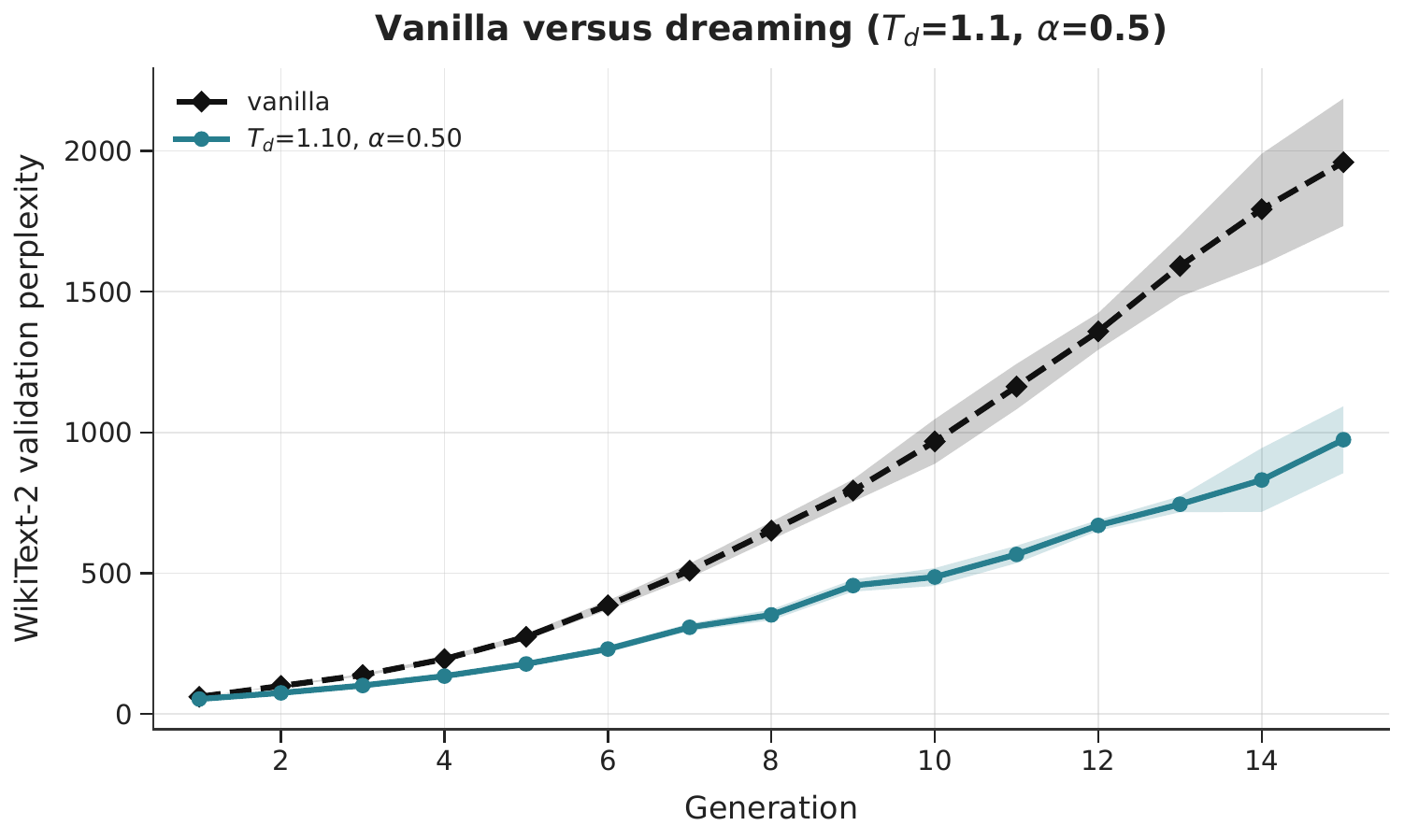}
\vspace{-0.2cm}
\caption{GPT-2 (124M) under the generate-and-retrain loop, with synthetic
corpora sampled at $T_s = 0.8$. WikiText-2 validation perplexity against 
generation count, vanilla and dreaming ($T_d = 1.1$, $\alpha = 0.5$), over 
five seeds; shaded regions represent 25th to 75th percentile.}
\label{fig:gpt2_ridgeline}
\end{figure}

We repeat the recursive protocol with a 4-layer LSTM language model \citep{hochreiter1997} trained on a small public-domain corpus (Appendix~\ref{app:lm}), monitoring validation loss as the number of self-generated generations grows, following \citet{gerstgrasser2024}. At $T_s=0.7$ the vanilla model's validation loss degrades with slope $8.7\cdot10^{-3}$ per generation, while at $T_s=1$ the slope is $9.6 \cdot 10^{-5}$: accumulation is close to harmless at unit temperature and clearly harmful below it (Figure~\ref{fig:lstm}). The dreaming model ($T_d=3$) is flat at both temperatures and, even at generation $0$, it already converges to a lower validation loss. 
As a scaling check we pre-trained GPT-2 (124M) \citep{radford2019} from
scratch on OpenWebText \citep{gokaslan2019}, fine-tuned on WikiText-2, and
iterated the generate-and-retrain loop with synthetic corpora sampled at
$T_s = 0.8$, holding out part of each synthetic corpus as a validation set. Figure~\ref{fig:gpt2_ridgeline} shows WikiText-2 validation perplexity across 15 generations. Both conditions degrade, but the dreaming run degrades more
slowly: at generation 15 the vanilla model reaches a perplexity of roughly
1960 against roughly 980 for $T_d = 1.1$, $\alpha = 0.5$, and the two
separate from about generation 5 onward with the ratio holding near a
factor of $2.0$ thereafter. Unlike Markov and LSTM, dreaming slows rather than arrests drift in GPT-2. This asymmetry warrants future investigation.

\section{Limitations}
\label{sec:limits}
A stable spectrum is not necessarily a correct one. The dreaming spectrum
in Figure~\ref{fig:spectrum} settles below the true subdominant
eigenvalues, so the learned chain mixes faster than the target. Second, we observe the temperature dependence at $T_s \in \{0.7, 0.8, 1.0\} $ across our three 
experimental settings, so the behaviour near $T_s = 1$ is suggested by the sharpening picture 
rather than established by a fine-grained sweep. Third, the argument of Section~\ref{sec:spectral}  treats the operator as exactly
recoverable, whereas this is not necessarily the case. Fourth, $T_d$ and $\alpha$ require empirical tuning (useful ranges $T_d \in [1, 3]$, $\alpha \in [0.3, 1]$, though optimal pairs vary by setting). At GPT-2 scale, tuning cost is the main obstacle to scaling, and dreaming roughly doubles per-step generation cost.
\section{Conclusion}
The eigenvalues of the learned transition operator give a view of model collapse that scalar loss does not. Under self-training at $T_s < 1$, subdominant eigenvalues drift toward the unit circle and the spectral gap narrows. Dreaming Learning leaves that spectrum flat, apparently by keeping operator entries away from zero through samples drawn at $T_d > 1$. The spectrum also reveals the method's cost: the learned chain mixes faster than the target. Open questions: whether residual bias can be reduced by annealing $T_d$, and whether the floor can be characterised to predict the optimum in $(T_d, \alpha)$.
\bibliographystyle{plainnat}
\bibliography{references}

\appendix

\section{Markov model implementation}
\label{app:markov}
\begin{figure}[h!]
\centering
\begin{subfigure}{0.48\linewidth}
  \includegraphics[width=\linewidth]{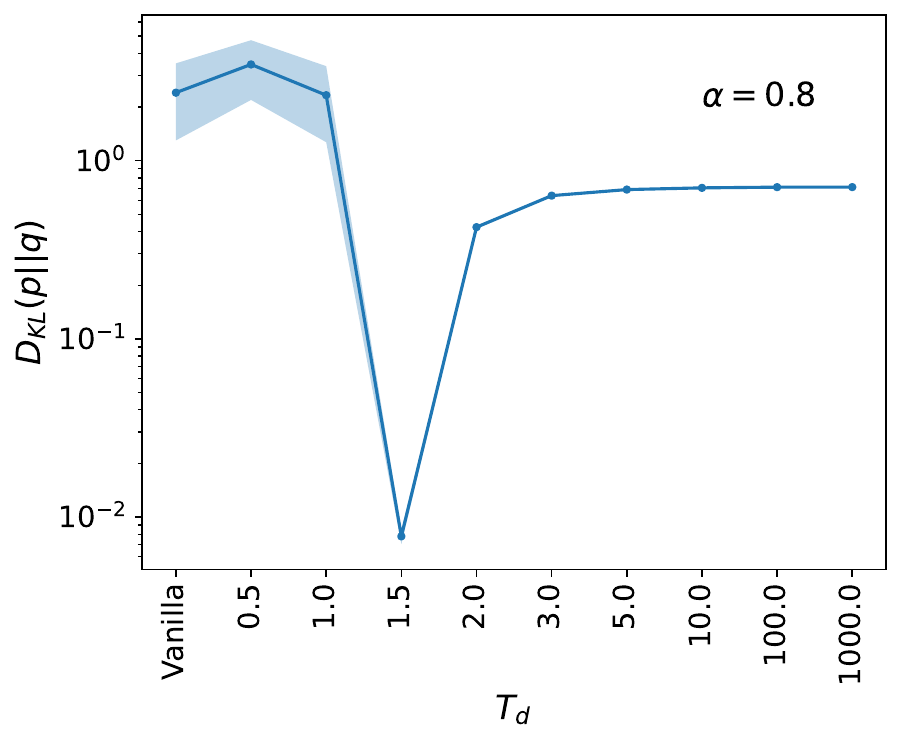}
\end{subfigure}\hfill
\begin{subfigure}{0.48\linewidth}
  \raisebox{0.55cm}{\includegraphics[width=\linewidth]{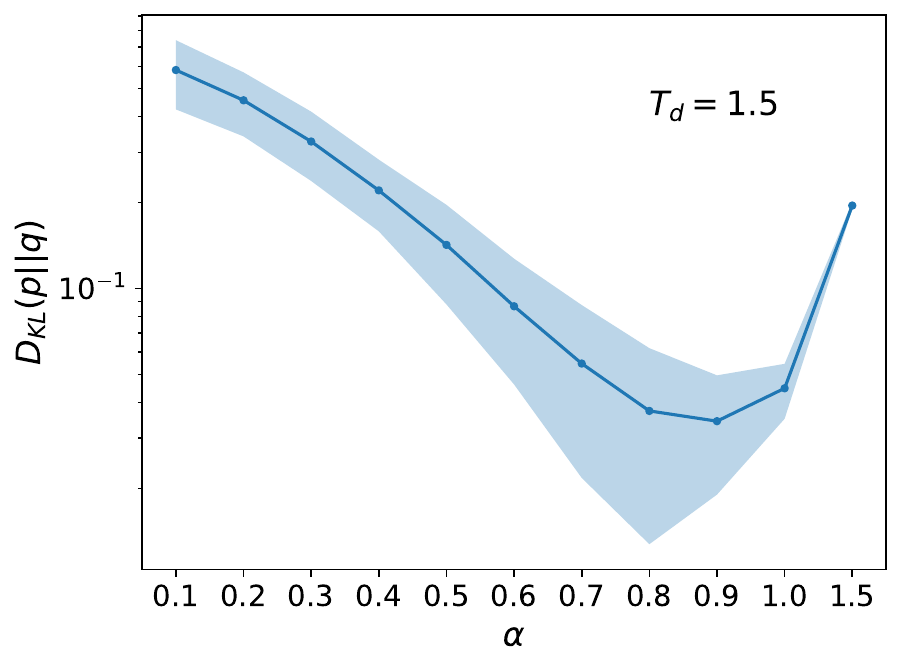}}
\end{subfigure}
\caption{$\KL(p\|q)$ between the true and reconstructed stationary
distributions after $5000$ generations over 10 seeds. \textbf{Left:} varying $T_d$ at
$\alpha = 0.8$. \textbf{Right:} varying $\alpha$ at $T_d = 1.5$.
\textsc{Vanilla} denotes no dreaming term. Shaded regions represent 25th to 75th percentile.}
\label{fig:kl}
\end{figure}

\begin{figure}[h]
\centering
\includegraphics[width=0.5\linewidth]{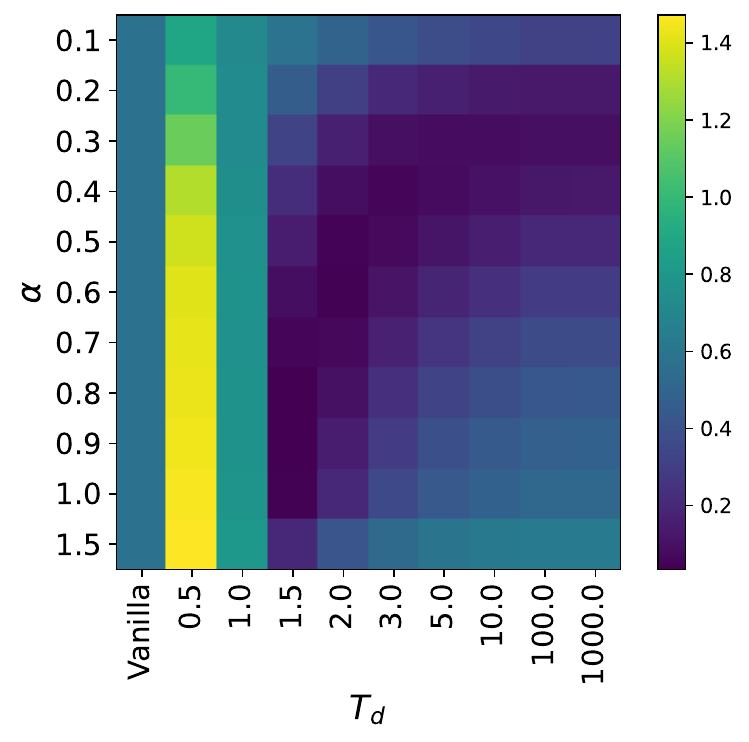}
\caption{$\KL$ between true and reconstructed stationary distributions
after $5000$ generations across the $(T_d, \alpha)$ grid.}
\label{fig:heatmap}
\end{figure}

$M_0$ has $n=10$ states and normalised stationary entropy $0.8$; results are unchanged for other entropies and state counts. The baseline dataset $S_0$ is a sequence of $20{,}000$ symbols sampled from $M_0$. At each generation, $N_g = 5000$ symbols are generated at $T_s = 0.7$ and concatenated, $S_{t+1} = S_t \oplus s_t$. The dreaming batch is $N_d = 5000$ symbols sampled recursively from the network's own tempered outputs at $T_d$, as described in Section~\ref{sec:dreaming}. The reconstruction uses a feed-forward network with two hidden layers of $50$ units and tanh activations, trained epoch-wise on the full accumulated dataset with reshuffling. Here $\alpha$ is the coefficient of the dreaming term in \eqref{eq:dreaming} and, being a loss coefficient rather than a mixing weight over data, is not restricted to $[0,1]$. All figures are over 10 seeds. Figure~\ref{fig:heatmap} shows the full $(T_d, \alpha)$ landscape and Figure~\ref{fig:kl} shows the respective slices demonstrating that optimal performance (lowest KL divergence) occurs at large $\alpha$, consistent with the mechanism of keeping operator entries away from zero.

\section{Language model implementation}
\label{app:lm}

The LSTM has 4 hidden layers of 400 units; input and output dimensions
equal the vocabulary size. Training texts are \emph{The Mysterious
Affair at Styles} and \emph{The Secret Adversary} (Christie) and
\emph{The Way We Live Now} (Trollope); validation uses \emph{Unbroken}
(Hillenbrand), \emph{Armadale} (Collins) and \emph{Desert Gold} (Grey).
All are public domain via Project Gutenberg. The vocabulary contains
$43{,}138$ word types and the training set $597{,}035$ tokens.
\subsection*{Meta-parameter selection for GPT-2}

\begin{figure}[t]
\centering
\begin{subfigure}{0.5\linewidth}
  \includegraphics[width=\linewidth]{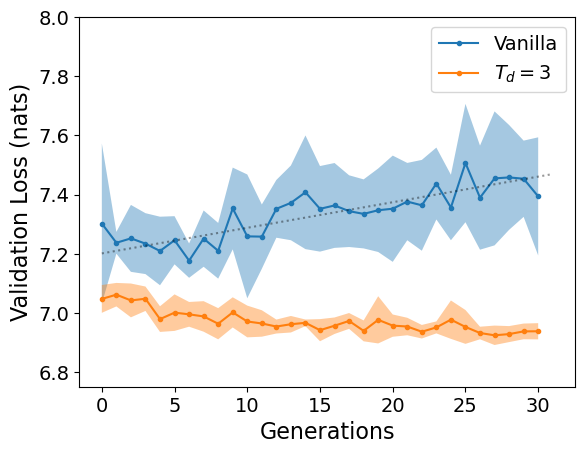}
\end{subfigure}\hfill
\begin{subfigure}{0.5\linewidth}
  \includegraphics[width=\linewidth]{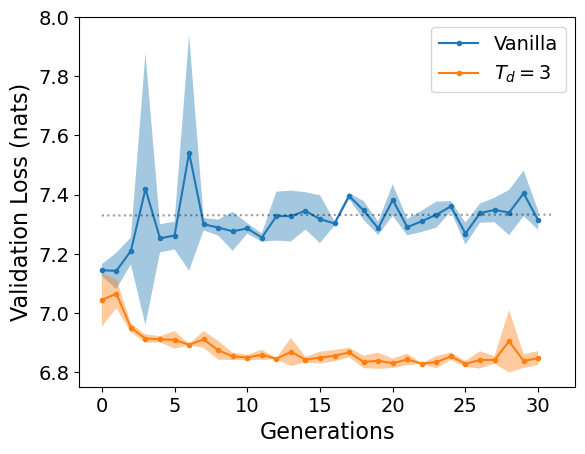}
\end{subfigure}\hfill
\caption{\textbf{Left, right:} LSTM validation loss against generation
count for vanilla and dreaming ($T_d{=}3$) models at $T_s{=}0.7$ and
$T_s{=}1$ over 10 seeds each. Shaded region represents 25th to 75th percentile.}
\label{fig:lstm}
\end{figure}

\begin{figure}[h]
\centering
  \includegraphics[width=0.8\linewidth, trim={0 0 0 27}, clip]{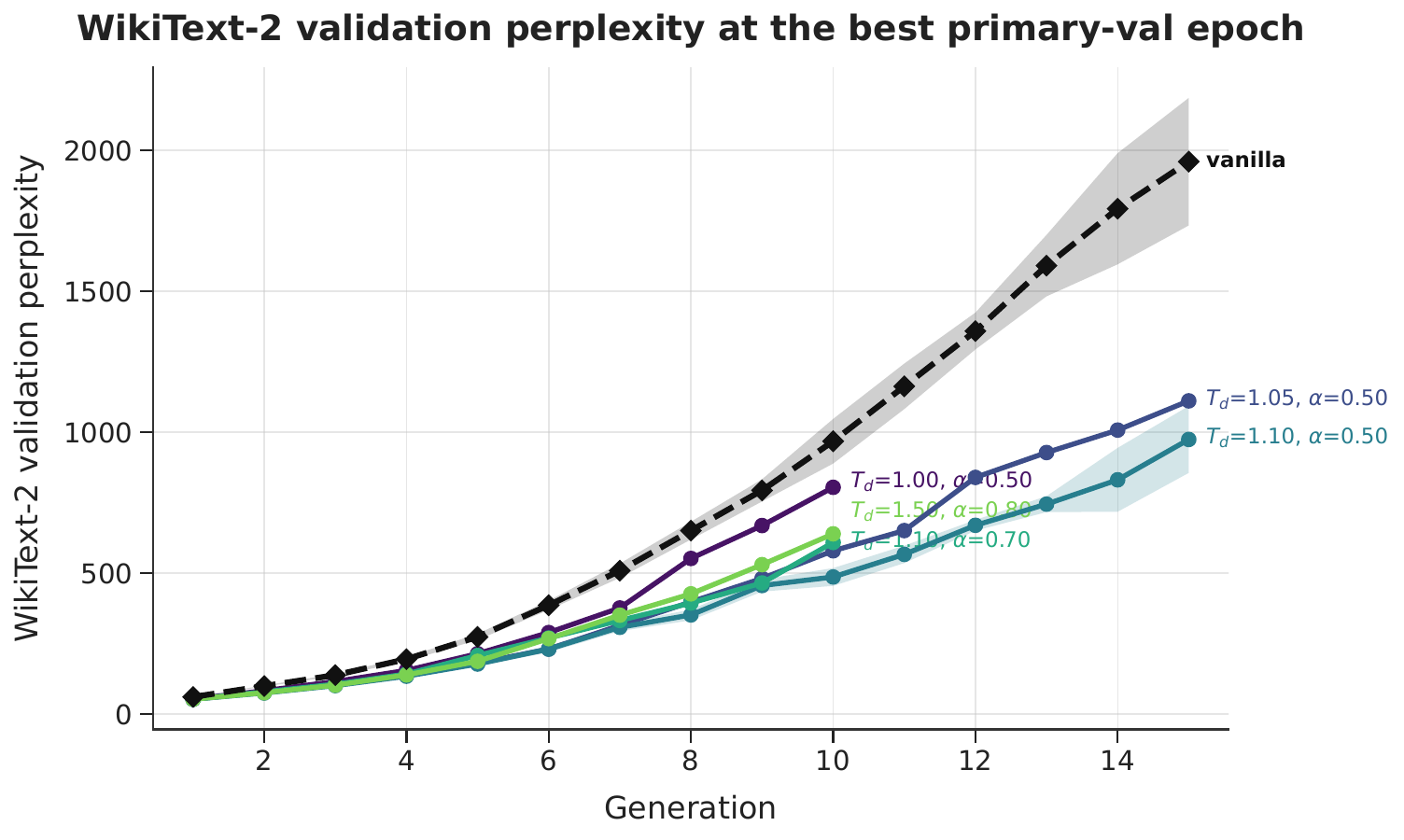}
\caption{WikiText-2 validation perplexity at the best primary-validation
epoch, for each $(T_d, \alpha)$ configuration tried. Only vanilla and $T_d = 1.1$, $\alpha = 0.5$ are averaged
over seeds, with shaded regions spanning the 25th to 75th percentile; the
remaining curves are single runs and some end at generation 10.}
\label{fig:sweep}
\end{figure}

A single trajectory at the GPT-2 scale is expensive: pre-training from scratch on OpenWebText requires approximately 42 hours on four A100 GPUs, fine-tuning on WikiText-2 takes roughly five minutes, and each of the 15 generate-and-retrain iterations takes approximately 10 minutes (synthesis plus retraining). A complete exhaustive grid over $T_d \in [1.0, 1.5]$ at intervals of $0.05$ and
$\alpha \in [0.3, 1.0]$ at intervals of $0.1$ would thus require $88$ trajectories,
or $88$ times the cost of a single run. The computational burden prompted a two-line sweep: fixed $\alpha = 0.5$ with $T_d \in \{1.00, 1.05, 1.10, 1.50\}$, and fixed $T_d$ values at $\alpha \in \{0.5, 0.7, 0.8\}$. This approach required approximately $36$ GPU-hours and was sufficient to locate a
configuration better than vanilla and to characterize the local landscape (Figure~\ref{fig:sweep}).

\end{document}